\documentclass{article}
\usepackage{float}
\usepackage[preprint]{neurips_2025}
\usepackage{comment}
\usepackage[utf8]{inputenc} % allow utf-8 input
\usepackage[T1]{fontenc}    % use 8-bit T1 fonts
\usepackage{hyperref}       % hyperlinks
\usepackage{url}            % simple URL typesetting
\usepackage{booktabs}       % professional-quality tables
\usepackage{amsfonts}       % blackboard math symbols
\usepackage{nicefrac}       % compact symbols for 1/2, etc.
\usepackage{microtype}      % microtypography
\usepackage{xcolor}         % colors
\usepackage{graphicx} % Required for inserting images

\title{OpenTwinMap: An Open-Source Digital Twin Generator for Urban Autonomous Driving}
\workshoptitle{Urban AI}
\author{%
  Alex Richardson \\
  Department of Computer Science\\
  Vanderbilt University\\
  Nashville, TN \\
  \texttt{william.a.richardson@vanderbilt.edu}
  \And
  Jonathan Sprinkle \\
  Department of Computer Science\\
  Vanderbilt University\\
  Nashville, TN \\
  \texttt{jonathan.sprinkle@vanderbilt.edu}
}
\date{September 2025}

\begin{document}

\maketitle

\begin{abstract}
Digital twins of urban environments play a critical role in advancing autonomous vehicle (AV) research by enabling simulation, validation, and integration with emerging generative world models. While existing tools have demonstrated value, many publicly available solutions are tightly coupled to specific simulators, difficult to extend, or introduce significant technical overhead. For example, CARLA-the most widely used open-source AV simulator-provides a digital twin framework implemented entirely as an Unreal Engine C++ plugin, limiting flexibility and rapid prototyping. In this work, we propose OpenTwinMap, an open-source, Python-based framework for generating high-fidelity 3D urban digital twins. The completed framework will ingest LiDAR scans and OpenStreetMap (OSM) data to produce semantically segmented static environment assets, including road networks, terrain, and urban structures, which can be exported into Unreal Engine for AV simulation. OpenTwinMap emphasizes extensibility and parallelization, lowering the barrier for researchers to adapt and scale the pipeline to diverse urban contexts. We describe the current capabilities of the OpenTwinMap, which includes preprocessing of OSM and LiDAR data, basic road mesh and terrain generation, and preliminary support for CARLA integration. 
\end{abstract}

\section{Introduction}

\subsection{Background}
Digital twins for 3d urban driving environments have become central to autonomous vehicle (AV) research \cite{zhao2024survey}. Among existing platforms, CARLA \cite{dosovitskiy2017carla} is the leading open-source 3D simulator, offering Unreal Engine extensibility, diverse vehicles, and a digital twin tool that generates maps from OSM and elevation data. However, CARLA's tool has various limits - it functions only as an Unreal plugin, has a limited asset library, and lacks support for bridges and overpasses, which are common in urban environments. 

Recent work has explored the use of generative models to increase realism by refining simulator sensor data, and enabling interactive scene editing \cite{zhao2024exploring, ren2025cosmos, gosselin2025ctrl, chen2024drivinggpt, gao2025foundation}. While promising, purely data-driven approaches such as CityDreamer \cite{xie2024citydreamer}, which builds 3D city models from Google Earth and OSM, cannot guarantee rule-based control or model complex structures such as tunnels. Similarly, world generation methods from sensor data \cite{ost2025lsd, bogdoll2025muvo, samak2025digital}, including prompt-driven environments \cite{samak2025digital} highlight the potential for flexible content creation but remain insufficient for research settings that demand reproducible, geometrically precise maps.

For AV researchers, there is a clear need for an efficient, realistic, flexible software framework for rule-based 3d urban environment digital twin map generation.

\subsection{Contributions of this Work}
We propose OpenTwinMap, an open-source framework for generating high-fidelity digital twins of urban environments tailored for autonomous driving research, with a flexible, parallelizable pipeline for road network and static environment object generation. 

\begin{enumerate} 
    \item \begin{bf} Open-source extensibility \end{bf}: An early in-development iteration of it is available on github under the GPL license: \url{https://tinyurl.com/ybvh43wr}. Currently it supports simple terrain generation and basic road mesh generation and placement.
    \item \begin{bf}CARLA integration along with flexibility for other simulators \end{bf}: Currently a CARLA stub plugin is also available \url{https://tinyurl.com/33sjzx6b} that directly loads the generated 3d assets and generates a CARLA map from them. Furthermore, the generic nature of the pipeline means it can be readily extended to other simulators.
    \item \begin{bf} \cite{opendrive} Python library  1.4 implementation with full modularity and an early customizable OSM-to-OpenDRIVE converter\end{bf}: We also provide an early implementation of the OpenDRIVE 1.4 specification that suits our use case. Furthermore, we use it to convert the OSM to OpenDRIVE, providing a full 3d description of the road network.
    \item \begin{bf}Parallelizable and extensible pipeline for 3D asset creation from OpenDRIVE and static environment object data\end{bf}: This allows for rapid prototyping and quick export of maps.
\end{enumerate}

\section{OpenTwinMap Framework}
\subsection{Overall}
\begin{figure}[H]
  \centering
  \label{architecture}
  \includegraphics[width=\linewidth]{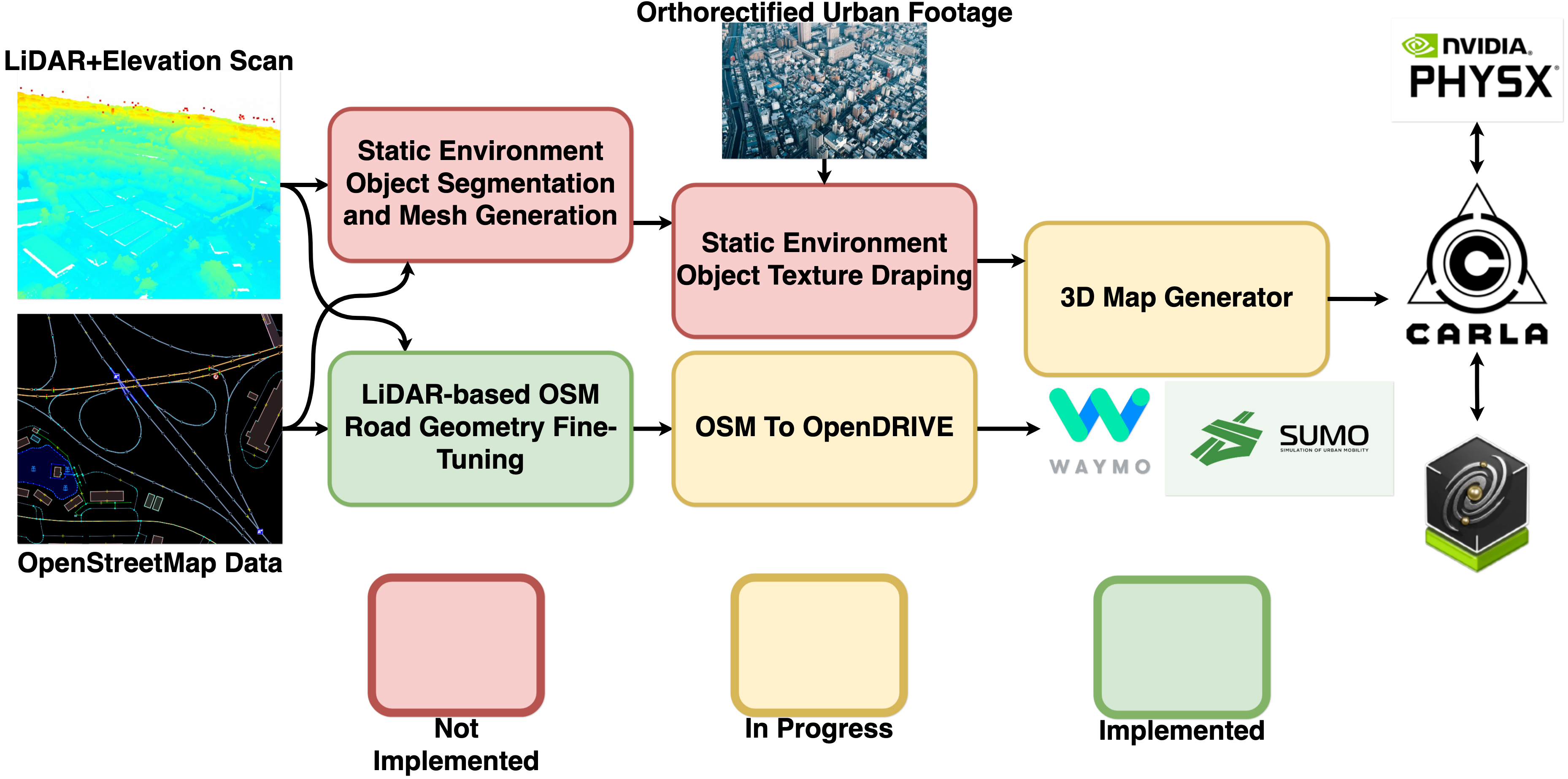}
  \caption{Architecture of OpenTwinMap outlined, with future features mapped together with what is already implemented or in progress. }
\end{figure}
The diagram describing the overall framework for OpenTwinMap is at Figure \ref{architecture}. As OpenTwinMap is currently in development, many aspects are yet to be completed. To construct high-fidelity urban digital twins, OpenTwinMap integrates multiple geospatial data sources through a modular pipeline. First, LiDAR-based OSM road geometry fine-tuning refines the inherently noisy road geometries in OpenStreetMap by aligning them with high-resolution LiDAR scans, producing geometrically consistent centerlines. Next, the OSM to OpenDRIVE stage transforms this into a fully compliant OpenDRIVE representation, enabling standardized descriptions of lanes, junctions, and traffic elements, and potential connectivity with SUMO and WayMax down the line.  Beyond the road network, static environment object generation will incorporate surrounding urban infrastructure, including buildings, barriers, and other immovable structures - ensuring semantic segmentation between different asset categories. Finally, the 3D map generator synthesizes these components into a set of meshes and textures to be imported into CARLA. Together, these stages will form an extensible and parallelizable workflow for end-to-end urban digital twin creation.

\subsection{Current Capabilities and Limitations}
Currently OpenTwinMap is in active development, so it's prudent to clearly distinguish between current capabilities and proposed-future capabilities:

Current Capabilities:
\begin{enumerate}
    \item Parallelized rapid OSM Road Geometry fine-tuning
    \item Basic road and lane OSM-to-OpenDRIVE conversion
    \item Road elevation integration that supports overpasses and basic bridges
    \item Modularized 3D mesh generation and Unreal Engine import system
\end{enumerate}

Proposed-Future capabilities:
\begin{enumerate}
    \item OSM to OpenDRIVE junction detection and conversion
    \item Arbitrary Static Environment Object Segmentation and Mesh Generation via LiDAR and OSM data
    \item Static Environment Object texture draping
    \item Integration into WayMax and SUMO using OpenDRIVE and vehicle trajectory data
\end{enumerate}

\subsection{LiDAR-based OSM Road Geometry Fine-Tuning}
OpenStreetMap provides a valuable starting point for road network data, but its geometric accuracy is often insufficient for high-fidelity simulation, especially in urban environments where lane alignment, curvature, and terrain conformance are critical. To address this, OpenTwinMap incorporates a fine-tuning stage that aligns OSM-derived road geometries with LiDAR point clouds. The LiDAR data captures dense surface information, enabling correction of misaligned centerlines, conformity with varying terrain, and refinement of junction geometry. Our approach uses Open3D's point cloud registration library to adjust the road networks to maximize alignment with the LiDAR scans. This process reduces positional errors in OSM data and enforces terrain alignment, making it suitable for downstream OpenDRIVE conversion and 3D mesh generation.
\subsection{OSM To OpenDRIVE}
Although OSM provides rich coverage of urban road networks, its lane-level description and junction geometries are often incomplete or ambiguous. Converting these datasets into the OpenDRIVE standard for AV research therefore requires careful resolution of such ambiguities. To address this, OpenTwinMap introduces an OSM to OpenDRIVE conversion module implemented in a modular Python style, supported by a preliminary OpenDRIVE library. The framework emphasizes extensibility, enabling researchers to customize conversion rules, lane configurations, and more to suit their needs. In contrast, CARLA's built-in converter is tightly coupled to Unreal Engine and offers limited flexibility for customization. By enabling iterative refinement of lane semantics and junction structures, OpenTwinMap provides a more adaptable workflow. The current implementation focuses on U.S. cities, and ongoing development is extending support for full junction geometries and broader geographic applicability. 
\subsection{Static Environment Object Segmentation and Mesh Generation}
Beyond the road network itself, realistic 3D urban AV digital twin maps require accurate representation of surrounding static infrastructure such as buildings, barriers, and other immovable urban objects. In OpenTwinMap, we propose as a future feature to combine OSM annotations with LiDAR-derived geometry to enable semantic segmentation of the urban scene. This process will ensure that simulation maps are not only visually and structurally consistent, but capable of supplying pre-labeled data for AV Perception and Planning ML modules, and generative models for enhancing simulator sensor data. Each type of OSM object, along with the method for processing the corresponding LiDAR data into a 3D mesh, will be modularized into classes. This will allow new categories of infrastructure to be introduced with minimal overhead, supporting customization to diverse cities and use cases.
\section{Static Environment Object Texture Draping}
A corresponding important direction for future development is the integration of texture draping for static environment objects. The surface appearance of these 3D objects is essential for a proper digital twin. A tried-and-true approach is to support the projection of aerial imagery, street-level photographs, and dashcam videos onto existing meshes, enabling visually realistic surfaces. The design goal is to keep the texture draping system modular and source-agnostic, allowing researchers to experiment with different imagery datasets and texture mapping techniques. 
\subsection{3D Map Generator}
The final stage of the pipeline synthesizes the OpenDRIVE road network and static environment objects into a unified 3D representation suitable for autonomous driving simulators. The current 3D map generator constructs Unreal Engine-compatible meshes for basic road meshes and terrain, with the ultimate goal to comprehensively generate full road and junction meshes, terrain and infrastructure. Furthermore, it is fully parallelized, allowing rapid asset creation, prototyping, and Unreal Engine import. 

\section{Conclusions and Future Work}
In this work, we introduced OpenTwinMap, an open-source framework for generating digital twins of urban environments to support autonomous driving research. The framework integrates OSM and LiDAR data to produce accurate and semantically segmented road networks and 3D meshes that are directly compatible with CARLA, with easy extensibility for other simulators. By emphasizing modularity and extensibility, OpenTwinMap lowers the barrier to producing high-fidelity urban environments.

Looking ahead, several extensions are planned for the framework. These include full support for, more detailed infrastructure assets for bridges and overpasses, and proposed modules for creating arbitrary static environment objects to flesh out map generation. The framework is still in its early phase and we invite feedback and different ways of approaching those future features.

\section{Acknowledgments}
This work is supported by the NSF GRFP program under award number 2444112, and we would also like to thank the Tennessee Department of Transportation for their LiDAR and elevation scan data of the state - it has been instrumental in our work thus far.
\bibliography{refs}

@inproceedings{dosovitskiy2017carla,
  title={CARLA: An open urban driving simulator},
  author={Dosovitskiy, Alexey and Ros, German and Codevilla, Felipe and Lopez, Antonio and Koltun, Vladlen},
  booktitle={Conference on robot learning},
  pages={1--16},
  year={2017},
  organization={PMLR}
}

@article{zhao2024survey,
  title={A survey on recent advancements in autonomous driving using deep reinforcement learning: Applications, challenges, and solutions},
  author={Zhao, Rui and Li, Yun and Fan, Yuze and Gao, Fei and Tsukada, Manabu and Gao, Zhenhai},
  journal={IEEE Transactions on Intelligent Transportation Systems},
  year={2024},
  publisher={IEEE}
}

@inproceedings{zhao2024exploring,
  title={Exploring generative ai for sim2real in driving data synthesis},
  author={Zhao, Haonan and Wang, Yiting and Bashford-Rogers, Thomas and Donzella, Valentina and Debattista, Kurt},
  booktitle={2024 IEEE Intelligent Vehicles Symposium (IV)},
  pages={3071--3077},
  year={2024},
  organization={IEEE}
}

@article{ren2025cosmos,
  title={Cosmos-Drive-Dreams: Scalable Synthetic Driving Data Generation with World Foundation Models},
  author={Ren, Xuanchi and Lu, Yifan and Cao, Tianshi and Gao, Ruiyuan and Huang, Shengyu and Sabour, Amirmojtaba and Shen, Tianchang and Pfaff, Tobias and Wu, Jay Zhangjie and Chen, Runjian and others},
  journal={arXiv preprint arXiv:2506.09042},
  year={2025}
}

@article{gosselin2025ctrl,
  title={Ctrl-Crash: Controllable Diffusion for Realistic Car Crashes},
  author={Gosselin, Anthony and Luo, Ge Ya and Lara, Luis and Golemo, Florian and Nowrouzezahrai, Derek and Paull, Liam and Jolicoeur-Martineau, Alexia and Pal, Christopher},
  journal={arXiv preprint arXiv:2506.00227},
  year={2025}
}

@article{chen2024drivinggpt,
  title={Drivinggpt: Unifying driving world modeling and planning with multi-modal autoregressive transformers},
  author={Chen, Yuntao and Wang, Yuqi and Zhang, Zhaoxiang},
  journal={arXiv preprint arXiv:2412.18607},
  year={2024}
}

@article{gao2025foundation,
  title={Foundation Models in Autonomous Driving: A Survey on Scenario Generation and Scenario Analysis},
  author={Gao, Yuan and Piccinini, Mattia and Zhang, Yuchen and Wang, Dingrui and Moller, Korbinian and Brusnicki, Roberto and Zarrouki, Baha and Gambi, Alessio and Totz, Jan Frederik and Storms, Kai and others},
  journal={arXiv preprint arXiv:2506.11526},
  year={2025}
}

@inproceedings{xie2024citydreamer,
  title={Citydreamer: Compositional generative model of unbounded 3d cities},
  author={Xie, Haozhe and Chen, Zhaoxi and Hong, Fangzhou and Liu, Ziwei},
  booktitle={Proceedings of the IEEE/CVF conference on computer vision and pattern recognition},
  pages={9666--9675},
  year={2024}
}

@article{ost2025lsd,
  title={LSD-3D: Large-Scale 3D Driving Scene Generation with Geometry Grounding},
  author={Ost, Julian and Ramazzina, Andrea and Joshi, Amogh and B{\"o}mer, Maximilian and Bijelic, Mario and Heide, Felix},
  journal={arXiv preprint arXiv:2508.19204},
  year={2025}
}

@inproceedings{bogdoll2025muvo,
  title={Muvo: A multimodal generative world model for autonomous driving with geometric representations},
  author={Bogdoll, Daniel and Yang, Yitian and Joseph, Tim and Yazgan, Melih and Zollner, J Marius},
  booktitle={2025 IEEE Intelligent Vehicles Symposium (IV)},
  pages={2243--2250},
  year={2025},
  organization={IEEE}
}

@article{samak2025digital,
  title={When Digital Twins Meet Large Language Models: Realistic, Interactive, and Editable Simulation for Autonomous Driving},
  author={Samak, Tanmay Vilas and Samak, Chinmay Vilas and Li, Bing and Krovi, Venkat},
  journal={arXiv preprint arXiv:2507.00319},
  year={2025}
}

@misc{opendrive,
	author = {OpenDRIVE},
	title = {opendrive.org},
	howpublished = {\url{http://www.opendrive.org}},
	year = {},
	note = {[Accessed 02-09-2025]},
}

\end{document}